\documentclass{article}

\usepackage{microtype}
\usepackage{graphicx}
\usepackage{subfigure}
\usepackage{booktabs} %
\usepackage{multirow}
\usepackage{amsmath}
\usepackage{siunitx}

\usepackage{hyperref}

\usepackage[accepted]{mlsys2025}
\usepackage{comment}
\usepackage{enumitem}
\usepackage{threeparttable} %

\usepackage[normalem]{ulem}

\mlsystitlerunning{Lightweight Software Kernels and Hardware Extensions for Efficient Sparse Deep Neural Networks on Microcontrollers}

\begin{document}

\twocolumn[
\mlsystitle{Lightweight Software Kernels and Hardware Extensions for Efficient Sparse Deep Neural Networks on Microcontrollers}

\mlsyssetsymbol{equal}{*}

\begin{mlsysauthorlist}
\mlsysauthor{Francesco Daghero}{to}
\mlsysauthor{Daniele Jahier Pagliari}{to}
\mlsysauthor{Francesco Conti}{bo}
\mlsysauthor{Luca Benini}{bo,zu}
\mlsysauthor{Massimo Poncino}{to}
\mlsysauthor{Alessio Burrello}{to}
\end{mlsysauthorlist}

\mlsysaffiliation{to}{Politecnico di Torino, Turin, Italy}
\mlsysaffiliation{bo}{University of Bologna, Bologna, Italy}
\mlsysaffiliation{zu}{ETH Zurich, Zurich, Switzerland}

\mlsyscorrespondingauthor{Francesco Daghero}{francesco.daghero@polito.it}

\mlsyskeywords{Deep Neural Network Deployment, TinyML, Pruning, Microcontrollers, RISC-V}

\vskip 0.3in

\begin{abstract}
The acceleration of pruned Deep Neural Networks (DNNs) on edge devices such as Microcontrollers (MCUs) is a challenging task, given the tight area- and power-constraints of these devices.
In this work, we propose a three-fold contribution to address this problem. First, we design a set of optimized software kernels for N:M pruned layers, targeting ultra-low-power, multicore RISC-V MCUs, which are up to 2.1$\times$ and 3.4$\times$ faster than their dense counterparts at 1:8 and 1:16 sparsity, respectively. Then, we implement a lightweight Instruction-Set Architecture (ISA) extension to accelerate the indirect load and non-zero indices decompression operations required by our kernels, obtaining up to 1.9$\times$ extra speedup, at the cost of a 5\% area overhead. 
Lastly, we extend an open-source DNN compiler to utilize our sparse kernels for complete networks, showing speedups of 3.21$\times$ and 1.81$\times$ on a ResNet18 and a Vision  Transformer (ViT), with less than 1.5\% accuracy drop compared to a dense baseline.

\end{abstract}
]

\printAffiliationsAndNotice{} %

\section{Introduction}\label{sec:intro}
The execution of Deep Neural Networks (DNNs) on extreme edge devices, such as IoT end-nodes based on Microcontrollers (MCUs), has become increasingly popular~\cite{wang2020deep}. Local execution enables smart capabilities in these devices while avoiding the costly transmission of raw data, with advantages in latency predictability, data privacy, and energy efficiency~\cite{sze2017efficient,shi2016edge}.
However, since edge devices operate on tight memory and power constraints, DNNs need to be extensively optimized before they can be deployed on MCUs. Techniques such as neural architecture search~\cite{fbnet}, quantization~\cite{integer_quantization,quantization_dl}, and pruning~\cite{scalpel,dcsr} aim at reducing DNNs' memory occupation and computational requirements while limiting accuracy drops.

In particular, weight pruning removes (i.e., sets to zero) the least relevant weights of a DNN, with potential benefits for both memory and computation,
as operations involving zeroed-out weights can be skipped~\cite{scalpel}.
However, sparse workloads have less regular memory accesses and lower arithmetic intensity than their dense counterparts, leading to lower-than-expected performance gains when leveraging HW/SW stacks not explicitly designed for sparsity, especially at low pruning ratios~\cite{scalpel}.
In the State-of-the-Art (SotA), this issue is tackled by a combination of model-, software- and hardware-level countermeasures. At the DNN model level, \textit{structured} or \textit{semi-structured} pruning forces specific patterns in the positions of non-zero (NZ) weights,
simplifying memory access and indices storage.
A popular example is N:M pruning, in which exactly N weights are NZ, in every group of M~\cite{nm_mask_training}.
Several solutions for accelerating sparse workloads have been proposed at lower levels of the stack, ranging from optimized software kernels to custom hardware. The latter includes complete accelerators~\cite{eyeriss2,demm}, functional units within a CPU pipeline~\cite{vegeta}, or tensor core extensions, such as the one featured by the NVIDIA A100 GPU family~\cite{tensorRT}.

Extending sparsity support to MCUs, however, is not trivial, as their tight area and power constraints often do not allow the additional overhead of a complete accelerator or complex core modifications.
The few existing works~\cite{sssr} target unstructured sparsity, not exploiting the advantages of constrained pruning patterns such as N:M. As a consequence, they require the availability of uncommon HW features, such as Streaming Semantic Registers (SSRs), which contribute significantly to the total area of the system.
At the same time, purely software solutions are complex to implement on MCUs, which feature simple ISAs with limited vectorial capabilities.

This work tries to overcome this gap by proposing a solution for the acceleration of semi-structured N:M sparsity on MCUs.
Specifically, our contributions are the following:
\begin{itemize}
\item  We design a set of software kernels for 1:4, 1:8, and 1:16 sparsity on MCUs, targeting both convolutions (Conv) and fully-connected (FC) layers. 
With software only, we achieve speedups ranging from 1.1$\times$ to 1.85$\times$ for Conv and from 1.02$\times$ to 3.4$\times$ for FC with respect to the best dense library available, on the RISC-V MCU presented in~\cite{vega}, featuring an 8-core Parallel-Ultra-Low-Power (PULP) cluster. 
\item We then extend the ISA of our target with a new, \textit{lightweight} instruction, aimed at improving the efficacy of our sparse kernels. 
Specifically, we accelerate the \textit{activations decimation} (i.e., the selection of activations corresponding to NZ weights), enabling efficient use of the available Single Instruction Multiple Data (SIMD) instructions. Implementing this extension incurs an area overhead of 5.0\%, but allows our kernels' to obtain up to 1.9$\times$ extra speedup.
\item Lastly, we instrument a DNN compiler~\cite{match} to use our sparse kernels. 
On two complete DNNs, a Convolutional Neural Network (CNN) and a Vision Transformer (ViT), we achieve a latency reduction of 1.31$\times$ and 1.43$\times$, respectively, using 1:4 sparsity, with no accuracy loss. At 1:16 sparsity, with an accuracy drop $<$1.5\%, we show speedups of 3.21$\times$ for the CNN and 1.81$\times$ for the ViT.
\end{itemize}

\section{Background}\label{sec:background}
\subsection{DNN Pruning}
Several works have shown how large portions of weights can be pruned with limited impact on the prediction quality of DNNs~\cite{llama_pruning,cnn_pruning}.
Optimized implementations of pruned models leverage compressed representations for the sparse weights matrices, storing only NZ elements and the corresponding coordinates.
Only the activations corresponding to NZ weights are loaded and processed at inference time, while computations associated with zeroed weights are skipped.

Our work focuses on optimizing the execution of an already pruned DNN. Therefore, it is orthogonal to the specific pruning \textit{strategy}, i.e., the method used to select what and when to prune (for example, using weights magnitude or 1st/2nd order loss approximations). We refer readers to~\cite{pruning_review} for a comprehensive survey on this topic. In the rest of this section, we focus instead on
the most common pruning \textit{patterns} (i.e., constraints on the position of NZ weights to improve sparse layers' efficiency); then, we detail common data structures for sparse tensors.
 
\paragraph{Pruning Patterns:}
\begin{figure*}[ht]
    \centering
    \includegraphics[width=\linewidth]{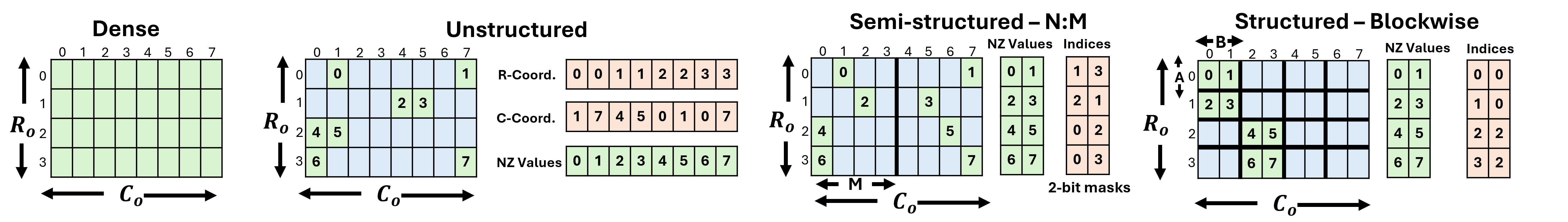}
    \vspace{-0.7cm}
    \caption{Pruning patterns and indices compressions with 75\% sparsity.}
    \label{fig:sparsity_pattern}
\end{figure*}

Fig.~\ref{fig:sparsity_pattern} depicts three common pruning patterns proposed in the literature.
\textit{Unstructured} (i.e., unconstrained) pruning leads to the highest compression ratio for a given accuracy target.
However, accelerating unconstrained sparse workloads is challenging: skipping computations requires performing multiple indirect memory loads with an irregular pattern, significantly impacting arithmetic density, as well as cache/scratchpad hit rates.
Therefore, for non-extreme sparsity ratios, layers with unstructured sparsity are often even slower than their dense counterparts~\cite{scalpel}.

\textit{Structured} pruning introduces constraints on the location of the NZ elements, yielding more regular data, at the cost of lower accuracy for the same sparsity ~\cite{blockwise,accelerating_vectorwise,accelerating_tilewise}.
For instance, \textit{block-wise sparsity}~\cite{blockwise} preserves dense groups of NZ elements of dimension $A \times B$ to increase L1 memory and register utilization,
but forcing such coarse-grained patterns usually yields large accuracy drops~\cite{blockwise}
Extremizing this concept, \textit{feature} or \textit{channel} pruning~\cite{pruning_channel_accel}, not shown in the figure, eliminates entire \textit{rows} of the weight matrix, producing a result that is equivalent to a dense (but smaller) weight tensor, thus removing all the difficulties of sparsity, but worsening performance even more.

The N:M approach represents a middle-ground between unstructured and block-wise sparsity (often referred to as \textit{semi-structured}), enforcing the constraint of N non-zero elements in every group of M values.
This scheme facilitates load balancing and parallelism at all levels (multicore, SIMD, etc), as equally-sized tensor portions comprising a multiple of M elements always require the same amount of computation, while also ensuring a partial locality in activation accesses.
NVIDIA GPUs have added hardware support for this format starting from the A100 series~\cite{a100gpu}, although limited to 2:4 or 1:2 patterns, depending on the data type.

\paragraph{Sparse Data Structures:}
Several formats have been introduced in the literature to store the NZ elements of a sparse tensor and the corresponding indices with different memory versus decoding overhead tradeoffs.
The simplest COOrdinates (COO) format stores a sparse matrix as three arrays, containing the NZ elements and their (row, column) position. While not requiring any extra computation to obtain the coordinates, this format has a non-negligible memory overhead:
using 8-bit integer values and 16-bit indices, the minimum sparsity required to balance the memory overhead is 75\%.
The Compressed Sparse Rows (CSR) format also uses three one-dimensional arrays containing the NZ elements, the number of elements per row, and the column indices.
In practice, CSR compresses the row indexes of the COO format, trading off speed for memory efficiency.
However, its memory overhead is still non-negligible, requiring more than 50\% sparsity to be advantageous on 8-bit quantized values.

The N:M format, shown in the center of Figure~\ref{fig:sparsity_pattern}, stores sparse data in two matrices.
The first contains the NZ values and has dimensions $(R_0,\frac{C_0}{M}N)$ . The second stores the relative indices of the NZ elements within each M-sized block, compressed in $\lceil log_2{(M)} \rceil$ bits, usually rounded to the nearest larger power-of-two (e.g., M=8 uses 4 bits).
This format enables memory-efficient storage even at low sparsity ratios, such as 1:2, but requires additional unpacking operations to extract the compressed indices.

\subsection{IoT Edge Nodes}
The compute platforms in IoT edge nodes are traditionally centered around low-power and low-cost MCUs. However, in recent years, these devices have been progressively equipped with more advanced features to support the efficient local execution of advanced applications such as Digital Signal Processing (DSP) and DNNs. Accordingly, it is not uncommon for modern MCUs to include multiple processing elements and specialized instructions such as SIMD. At the same time, since energy and area costs remain critical, these devices often give up cache memories in exchange for a multi-level hierarchy of software-controlled scratchpads, where data transfers between levels are handled through Direct Memory Access (DMA) controllers.
Companies like STM~\cite{stm32}, NXP~\cite{nxp4300} and GreenWaves~\cite{gap9} have already commercialized such architectures.

In this work, we target the Vega PULP SoC~\cite{vega}, which features 10 RISC-V cores and is the blueprint architecture for commercial products such as GreenWaves' GAP9~\cite{gap9}. In Vega, one core acts as the Fabric Controller (FC), managing peripherals and orchestrating the entire workload; one controls the DMA set-up; the other 8 are organized as a cluster to speed up computation. All cores feature an extended ISA including SIMD dot product instructions on 8-bit integers, loads with automatic post increment, and hardware loops to accelerate DSP and DNN workloads. The SoC features a 128 kB L1 data memory, shared between all cluster cores, a 1.6 MB L2 main memory (comprising an MRAM memory, which we do not exploit), and 16 MB of external L3 HyperRAM memory.

\section{Related Works}~\label{sec:related}
Many works in the literature have focused on accelerating sparse DNN workloads.
Some of them target high-end CPUs/GPUs, either introducing novel software kernels to exploit existing HW or proposing HW extensions to achieve higher speed-ups.  For instance, the authors of~\cite{sparse_dnn,venom,vegeta} exploit graph-level optimizations, custom sparsity formats and new instructions to achieve speedups of up to 4x w.r.t. dense baselines.
Other works design complete HW accelerators, achieving impressive speed-ups 
at the cost of significant area overheads~\cite{eyeriss2,demm,eie}.

While less explored, some articles also target the execution of sparse DNNs on MCUs.
In~\cite{scalpel}, the authors propose pruning groups of weights with the same dimension as the target's SIMD width, so that, at inference time, SIMD instructions can still be utilized. This is a particular instance of blockwise pruning with dimensions A $\times$ B = 1 $\times$ SIMD-width.
They benchmark their custom kernels on an ARM Cortex M4 MCU with 2-way fixed-point SIMD  instructions, obtaining maximum speedups of 1.38$\times$ on a ConvNet at 59.95\% sparsity and of 3.51$\times$ on a LeNet-5 at 93.28\% sparsity, among CNNs. Additionally, they achieve a speed-up of 9.17$\times$ at 93.07\% sparsity on a LeNet300, composed only of FC layers. However, 
on the tested MCU, weights are directly loaded from Flash memory, with high latency. Therefore, the benefits of skipped loads due to SIMD-aware pruning hide the overheads of blockwise sparse processing, ``idealizing'' the speedups (absolute inference times would remain very high and bottlenecked by memory accesses).
Moreover, as mentioned in Sec.~\ref{sec:background}, blockwise sparsity can lead to large accuracy drops on more complex DNNs and datasets~\cite{blockwise} (LeNet is tested on MNIST, while ConvNet reaches 81.86\% accuracy on CIFAR10).

The work of~\cite{dcsr} focuses on unstructured sparsity, proposing a linear encoding for CSR indices to reduce their memory overhead in exchange for extra decoding operations. 
They benchmark their approach on an ARM Cortex-M55, using three DS-CNN variants and comparing with two dense baselines derived from CMSIS-NN~\cite{cmsisnn}, as well as with the storage of relative indices (RI) in CSR.
They achieve a maximum reduction of cycles on the largest DS-CNN of 41.7\% compared to the original CMSIS-NN library and 6.7\% compared to hand-optimized kernels, with 90\% weights sparsity on point-wise Convolutions, and FC layers. 
They also show a 5.21$\times$ memory reduction on the pruned layers, with a memory footprint only 3.5\% higher than the CSR's RI storage. 
Speedups compared to optimized dense kernels are relatively low due to the large decoding overhead.

The authors of~\cite{indexmac} implement an ISA extension to accelerate N:M sparsity on a 64-bit RISC-V core paired with a 16-lane 512-bit vector engine.
Their new instruction, called IndexMAC, allows low-cost indirect access to row portions of the dense activation matrix, pre-loaded in a vector register file.
Moreover, they introduce a set of optimized kernels performing row-wise matrix multiplication with this instruction, achieving  1.82$\times$/2.14$\times$/1.92$\times$ speed-ups over a SW-only sparse baseline on ResNet50, Densenet121, and InceptionV3 at 75\% sparsity.
Differently from our work, the deployment target is a high-end RISC-V core, leading to significantly different design choices.
First, MCUs do not include a dedicated vector register file and have limited options for data pre-loading, given their small L1 memory size.
Second, in their work,  NZ indices are stored uncompressed, limiting the memory reduction in exchange for increased throughput.
Finally, their vector engine supports 32-bit element vectors and scalar-by-vector operations, while we target 8-bit integer SIMD without any scalar-by-vector instruction.

In~\cite{sssr}, the authors introduce a RISC-V ISA extension to accelerate computations with unstructured sparsity, supporting both CSR and Compressed Sparse Fiber (CSF) formats.
They extend RI5CY cores with Sparse Stream Semantic Registers (SSSRs), which transparently load/store data based on a stream of indices, making them available to other instructions (e.g., dot products). This increases FPU utilization by getting rid of explicit load/store instructions.
While extremely effective (5x speedup at 95.7\% sparsity on GEMMs ), SSSRs are targeted for floating point kernels and are not easily extensible to support low-precision integer SIMD calculations. Moreover, SSSRs are complex circuits whose area overhead ($\approx$20 to 31 kGE depending on the configuration) is already significant (20-31\%) when considering an FPU-equipped RI5CY with its 102 kGE~\cite{ssr}, as baseline. When compared to an FPU-less RI5CY, the overhead increases to as much as 44\%. 
Lastly, CSR and CSF formats require the storage of NZ indices at high precision, thus resulting in significantly higher memory overheads w.r.t. the N:M format.

\section{Lightweight Kernels for N:M Sparse DNNs on MCUs}~\label{sec:methods}
\begin{table}[t]
\centering
\scriptsize
\caption{Layer dimensions notation}\label{tab:notation}
\begin{tabular}{ccc}
\hline
\textbf{Tensors} & \textbf{Dimensions} & \textbf{Abbreviations}   \\\hline
Input (I) & rows, columns, channels  &  IX/IY/C \\
Output (O) & rows, columns, channels         &  OX/OY/K \\
Weights (W) & filter height/width, channels in/out                 & FX/FY/C/K  \\
Activations & Padding/Stride & P/S   \\\hline
\end{tabular}
\end{table}
As outlined in Section~\ref{sec:intro}, the novelty of our work lies in being the first, to the best of our knowledge, to propose a way to accelerate N:M sparsity on MCUs, either through SW or through an area-inexpensive HW extension, without relying on ad-hoc accelerators.
The aim is to unlock a new set of Pareto-optimal trade-offs in the space of model accuracy vs. latency reduction vs. area cost, bridging the gap between the significant speedups offered by dedicated HW accelerators and classic MCUs without HW and SW extensions.

This section describes our sparse convolutional (Sec~\ref{sec:conv}) and FC (Sec.~\ref{sec:fc}) kernels, focusing first on SW-only versions and then showing how those can be enhanced with a lightweight ISA extension. We detail the HW that implements our new instruction in Sec.~\ref{subsec:hw}.
Our kernels operate on 8-bit data, already stored in L1 memory. To execute full networks, we integrate them in the DNN compiler of \cite{match}, as described in Sec.~\ref{subsec:match}.
We focus on semi-structured N:M sparsity, given the demonstrated benefits over an unstructured approach with a limited loss in accuracy \cite{nm_mask_training}. Namely, we support 1:4, 1:8, and 1:16 formats, as increasing sparsity further would lead to a too high accuracy loss, while reducing it would not lead to any latency benefit.
While we target the SoC of~\cite{vega}, our SW-only kernels are general enough to be used on all MCUs from the PULP family, i.e., RISC-V-based multi-core MCUs with hierarchical scratchpad memories \cite{xpulpv2}, and are written in plain C code.
Table~\ref{tab:notation} reports the notation used in the rest of the paper for layer hyper-parameters.

In all our kernels, we use the format shown in Fig.~\ref{fig:sparsity_pattern} and detailed in Sec.~\ref{sec:background} to store N:M sparse weights. Namely, only NZ weights are stored, together with their indices within the corresponding M-sized block, compressed on 4-bits (for 1:8 and 1:16 sparsity) or 2-bits (for 1:4).
Thus, 1:4 sparsity leads to a weight memory reduction of 68.75\%, 1:8 of 81.25\%, and 1:16 of 90.62\% (with int8 weights).
Notice that, in the same conditions, the CSR format requires storing $K$ row pointers and $\frac{K \times FX \times FY \times C}{M}$ column indices with a minimum precision of 16-bit (for reasonably sized layers), leading to less than 25\% compression at 75\% sparsity (i.e., the equivalent of the 1:4 format).

\subsection{Convolutional Kernel Library}\label{sec:conv}
\subsubsection{Dense Baseline}
\begin{figure}[t]
    \centering
    \includegraphics[width=0.95\linewidth]{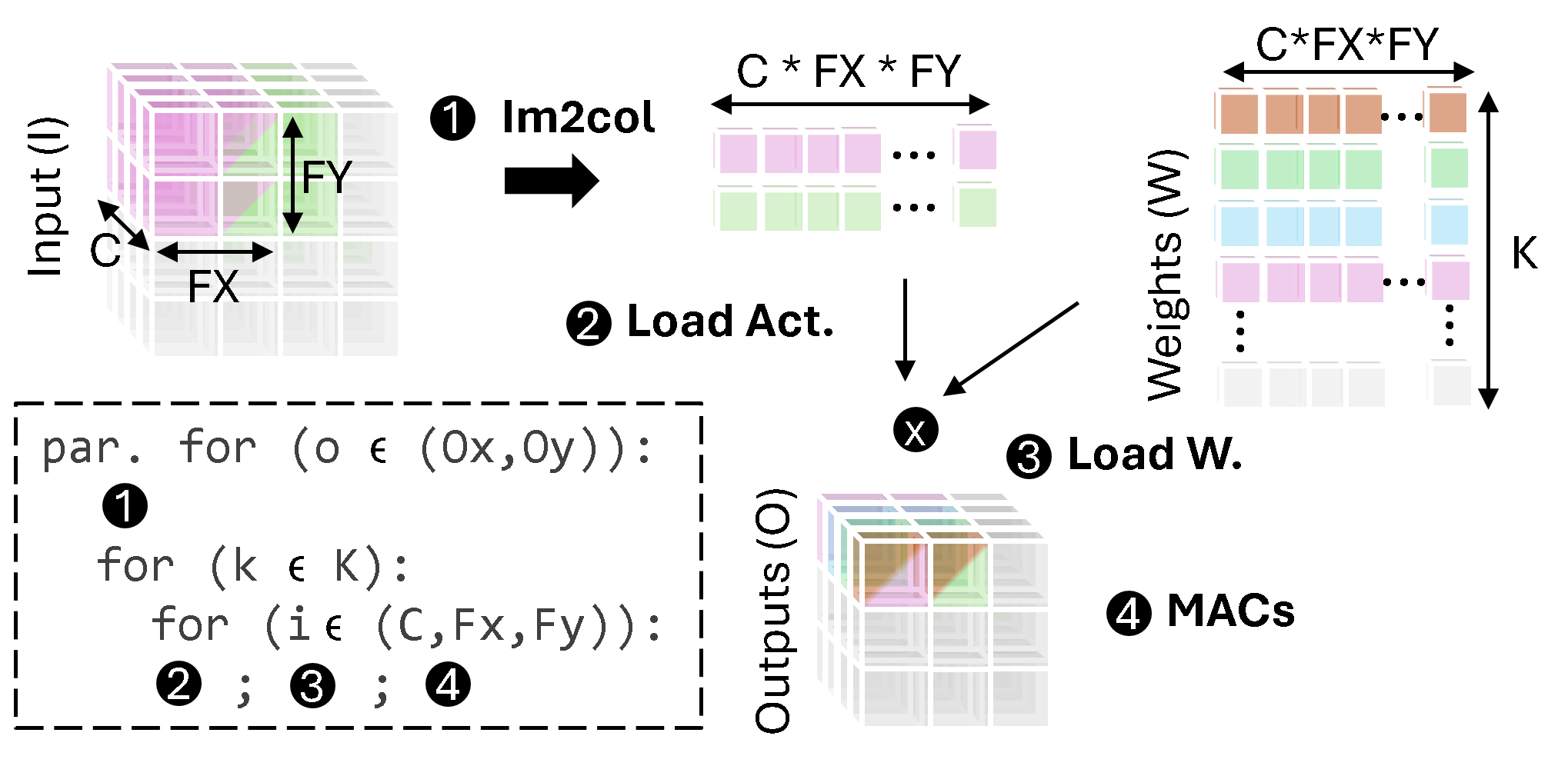}
    \vspace{-0.4cm}
    \caption{Inner loop of the PULP-NN dense convolutional kernel.}
    \label{fig:convolution_dense}
\end{figure}

The dense 8-bit kernels utilized as a baseline in our work are taken from the SotA PULP-NN library, described in \cite{pulpnn}.
Fig.~\ref{fig:convolution_dense} visualizes their internal loop: after performing a partial im2col transformation that reorganizes 2 spatially contiguous input patches into 1-dimensional arrays, the kernel loads the weights of 4 convolutional filters and iterates to produce a total of 8 outputs, (2 different spatial positions, and 4 output channels).
The dataflow is output-stationary; all output channels relative to the same spatial positions are produced before moving to the next ones. The detailed loop order is shown in the bottom-left of the figure. The outermost loops over OX and OY are parallelized on the platform's cores. Furthermore, the MACs of the innermost loop are performed using 4x8-bit SIMD dot product instructions available in the RISC-V XpulpV2 ISA extension~\cite{xpulpv2}. Notice that the partial im2col operation leads to an additional L1 memory requirement of FX$\times$FY$\times$C$\times$2$\times$N\_CORES elements.

This kernel achieves a theoretical peak performance in the innermost loop of 2.28 8-bit MACs/instruction/core (32 MACs with 14 instructions, i.e., six load words and 8 SIMD dot-products). However, it cannot be directly extended for sparse layers, as the inner loop unrolling assumes that the input activations used to generate all 4 output channels are identical, which enables reusing the im2col buffers. 
This is no longer true for N:M sparse layers.
Therefore, given the impossibility of implementing such unrolling for sparse kernels, we also consider a dense variant with 1x2 unrolling, i.e., computing a single output channel per internal loop iteration, that reaches a lower theoretical peak of 1.6 MACs/instruction/core.
The internal loop of this variant is shown on the left of Fig. \ref{fig:conv_code}.

\subsubsection{Software-only Sparse Kernel}
The SW-only versions of our sparse kernels only exploit instructions from the XpulpV2 extension, already available in~\cite{vega}. 
They have the same data flow and data stationarity of the dense baseline and use an innermost loop unrolling factor of 1x2, i.e., over two consecutive input patches and one weight filter. The single output channel per iteration is due to the reasons mentioned above.
Instead, further increasing the unrolling factor over the input patches would improve offsets and weights reuse but would lead to a linear increase in the im2col buffer memory overhead. This, in turn, would limit the usability of our kernels for layers with a large number of input channels or big filters, given that the im2col array must necessarily fit in L1.

The key difficulty in implementing kernels for N:M pruning is optimizing the loading of input activations corresponding to NZ weights in the innermost loop. We explored three alternatives:

1) \textit{DMA-based copy}: modify the DMA calls used to move data from L2 to L1 memory
to move, 
for each output channel, only activations corresponding to NZ weights (bypassing the im2col entirely).
However, this would eliminate the advantages of DMA burst transfers, leading to highly inefficient loads.

2) \textit{Sparse Im2col}: modify the im2col step to fill the 1-d buffers only with activations corresponding to NZ weights. However, there would be no reuse opportunity for these buffers, as different output channels do not share the NZ indices. The im2col would therefore become a part of the innermost loop, being repeated for every output channel, with a consequent explosion of innermost loop instructions.

3) \textit{Decimate Im2col}: keep the im2col step unchanged while adding a \textit{decimate} step in the innermost loop, that selects only the elements corresponding to NZ weights from the im2col buffer, for each different output channel. Compared to the previous option, the main difference is that the addresses of activations corresponding to NZ weights have to be computed relative to the im2col buffer, and not to the original input tensor. 
In other words, they are computed as \texttt{i$\times$M + o}, where \texttt{i} identifies the correct block of M elements in the im2col buffer, and \texttt{o} is the relative offset of the NZ weight within the block.
Conversely, to load activations directly from the input tensor, we should consider its spatial size (IY, IX) and the center position of each patch to compute the correct address, considering corner cases for different stride and padding combinations, leading to a high number of indexing operations in the innermost loop.

\begin{figure}[t]
    \centering
    \includegraphics[width=\columnwidth]{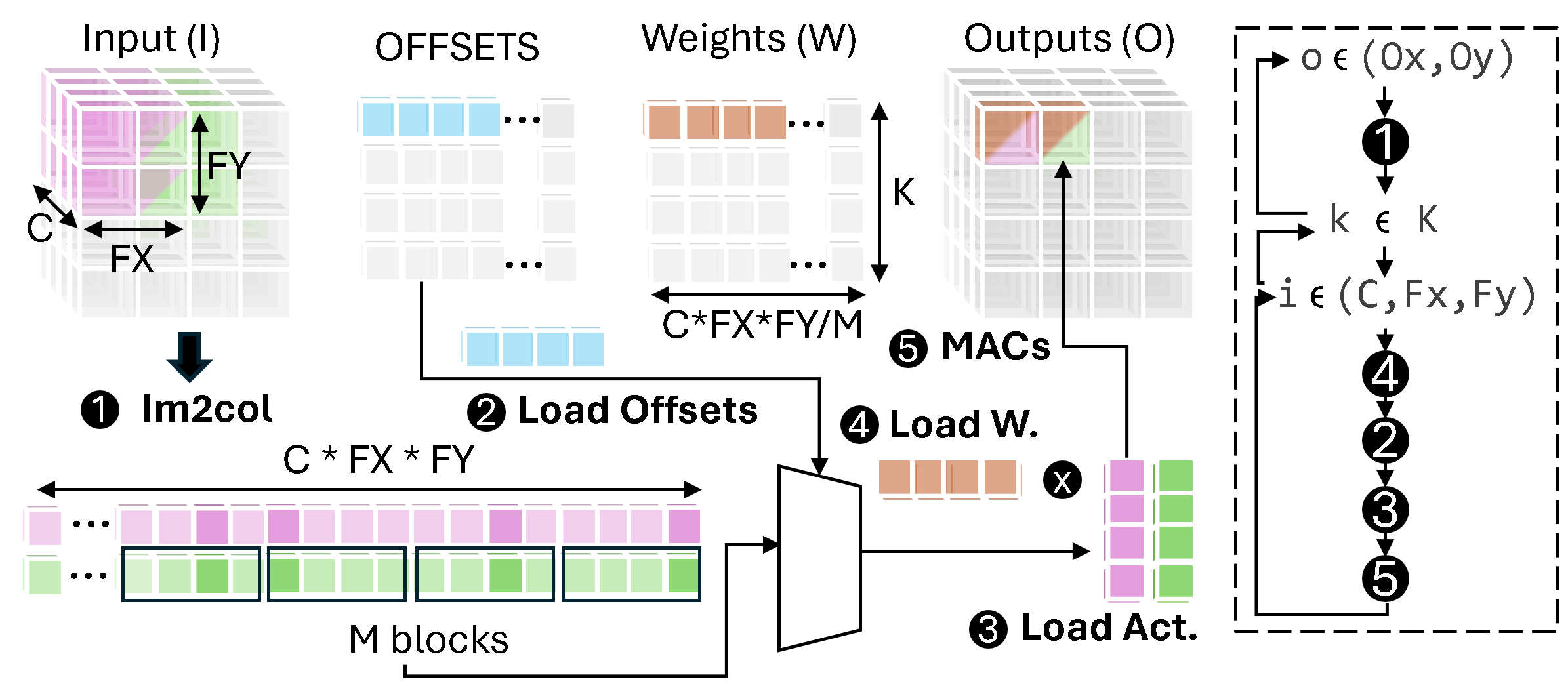}
    \vspace{-0.5cm}
    \caption{Inner loop of our sparse convolutional kernel. }
    \label{fig:software_sparse_convolution}
\end{figure}
\begin{figure*}[t]
    \centering
    \includegraphics[width=1\linewidth]{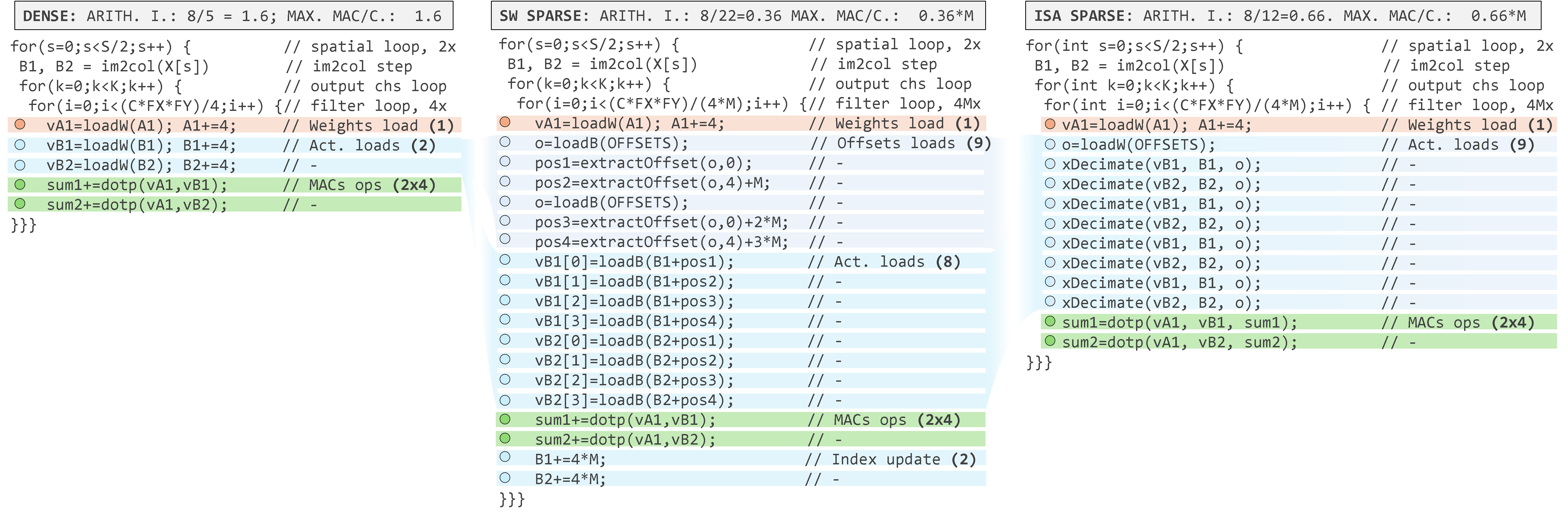}
    \vspace{-0.6cm}
    \caption{Innermost iteration of the dense matmul kernel (left), 1:8 / 1:16 sparse kernel with no custom instructions (center), and 1:8 / 1:16 sparse kernel with the \texttt{xDecimate} instruction (right).}
    \label{fig:conv_code}
\end{figure*}
Fig.~\ref{fig:software_sparse_convolution} visualizes our sparse convolutions, which use the latter strategy (\textit{Decimate Im2col}).  Fig.~\ref{fig:conv_code} (center) details the corresponding innermost loop execution.
Here, \texttt{OFFSETS} is the array whose elements store the relative indices of NZ elements in each M-sized block, on 4-bit (for 1:8 and 1:16 sparsity) or 2-bit (for 1:4);
the \texttt{extractOffset} function uses shift and mask operations
to unpack the correct index from each byte-sized element of the array.
\texttt{vB1} and \texttt{vB2}, the two 32-bit registers that store input activations, are filled using the four unpacked indices \texttt{pos<i>}.
Weights loading and dot products are identical to the dense baseline.

Each iteration of the 1:8 and 1:16 kernels performs 8 MACs and 22 instructions, leading to a peak performance of 0.36 MACs/instructions/core.
The 1:4 version has all the four indices required to fill a 32-bit word stored in a single 8-bit \texttt{OFFSETS} element, but requires more complex unpacking. Overall, it requires 23 instructions (2 more maskings, one less load) per iteration, leading to a peak of 0.35 MACs/instruction/core. 
Considering the number of theoretical dense MACs executed, i.e., multiplying the effective number of MACs by the sparsity factor M, we obtain peak performances of 1.4, 2.88, and 5.76 MACs/instruction/core. 

Notably, although our implementation is partly target-specific, the same design choices performed for our kernels can be extended to support other MCUs, without relying on features like on-board DMA or the XPulpV2 ISA extension.
For instance, ARM Cortex-M architectures leverage CMSIS-NN~\cite{cmsisnn} kernels and SIMD operations for dense convolutions, thus very similar kernel routines can be designed to add support for N:M sparsity.

\begin{figure*}[ht]
    \centering    \includegraphics[width=1.0\linewidth]{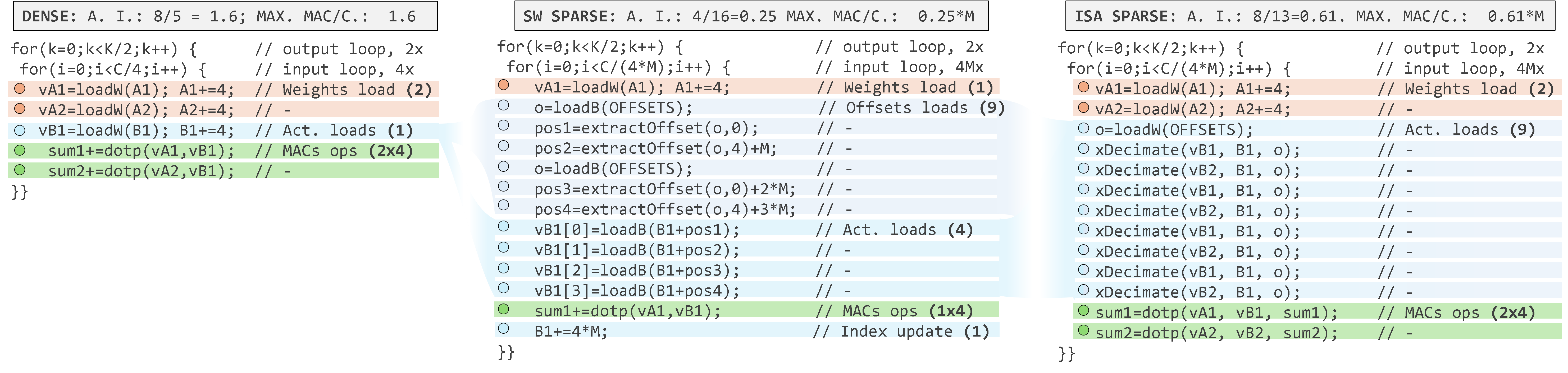}
    \vspace{-0.65cm}
    \caption{Innermost iteration of the dense FC kernel (left), 1:8 SW-only sparse kernel (center), and 1:8 ISA-extended sparse kernel (right).}
    \vspace{-0.6cm}
    \label{fig:fc_code}
\end{figure*}
\subsubsection{ISA-extended Sparse Kernel}
\label{subsec:conv_isa}
The main bottleneck of the SW-based sparse kernels lies in the indices unpacking and vector packing operations; 19 instructions are required to load the two activation registers, \texttt{vB1} and \texttt{vB2} (9 for computing indices, 8 for loading data, 2 for updating addresses).
To tackle this limitation, we design a new instruction, \texttt{xDecimate}, to load an 8-bit element from a buffer into a register, given an address and an offset, essentially merging the $\texttt{extractOffset}$ function and the load byte operation of the SW-only kernel.
Thanks to it, we reduce the instructions to fill the input registers to 8, and the total instructions in the innermost loop from 22 (or 23) to 12, regardless of the sparsity level, as shown in Fig.~\ref{fig:conv_code} (right).
Thus, we reach a peak of 0.66 MACs/instruction/core, corresponding to 2.64, 5.28, and 10.56 equivalent dense MACs/intruction/core, respectively at 1:4, 1:8 and 1:16 sparsity.

The syntax of the instruction is the following: \texttt{xdecimate rd, rs1, rs2}, where \texttt{rd} is the register in which extracted data will be stored, while the two source registers contain the starting address of the im2col buffer (\texttt{rs1}) and the packed NZ offsets (\texttt{rs2}).
At a high level, each execution of the instruction loads one activation byte in \texttt{rd}, computing the starting address of the target M-sized block in the im2col buffer, and adding the correct offset to it. It does so by combining \texttt{rs1} and \texttt{rs2}, with the value stored in one control-status-registers (csr\footnote{We use a lowercase acronym to avoid confusion with CSR, the Compressed Sparse Row format.}). The csr is auto-incremented at every \texttt{xDecimate} operation so that consecutive calls to \texttt{xDecimate} automatically point to the correct data. A specific instruction, \texttt{xDecimate.clear}, is used to reset it to 0 at the end of the loop over the output channels.

To account for the innermost loop unrolling (over two im2col buffers), the M-sized block's base address and the offset in the destination register are updated only once every two \texttt{xDecimate} executions. Instead, the \texttt{rs2} bits from which the NZ offset is unpacked are updated every time. This requires \textit{duplicating each NZ index} in the \texttt{OFFSETS} array, with a consequent memory overhead. However, it allows us to accommodate, with a single instruction, also the case of FC layers, as detailed in Sec.~\ref{sec:fc_isa}.
Notably, the overhead is acceptable given the low bit-width of indices; we still obtain weight memory savings of 62.5\% for 1:4 sparsity, 75\% for 1:8, and 87.5\% for 1:16.
The hardware implementation of \texttt{xDecimate} is detailed in Sec.~\ref{subsec:hw}.

\subsection{Fully-Connected Kernel Library}\label{sec:fc}
\subsubsection{Dense Baseline}
The innermost loop of our dense baseline for FC layers (again taken from PULP-NN), is shown in Fig.~\ref{fig:fc_code} (left). It is unrolled by a factor of 2 over the K dimension, as FC layers have no opportunities for weight reuse. The efficiency peak for this kernel is 1.6 MACs/instruction/core.
Multi-core parallelization is performed on the K dimension.

\subsubsection{Software-only Sparse Kernel}
We use the same innermost loop as for convolutions, unpacking the indices of four NZ elements and then using them for dot products. The pseudo-code is shown in Fig.~\ref{fig:fc_code} (center).
Differently from the dense baseline, we do not unroll over two output channels, given that each of them requires different input data. The theoretical peak performance is 0.25 MACs/instruction/core. Considering the equivalent dense MACs, we obtain 1.0, 2.0, and 4.0 MACs/instruction/core for 1:4, 1:8, and 1:16 sparsity. Notice that the 1:4 SW-only sparse kernel does not even reach sufficient theoretical performance to outperform the dense baseline.

\subsubsection{ISA-extended Sparse Kernel}\label{sec:fc_isa}
\begin{figure}[t]
    \centering
    \includegraphics[width=0.9\linewidth]{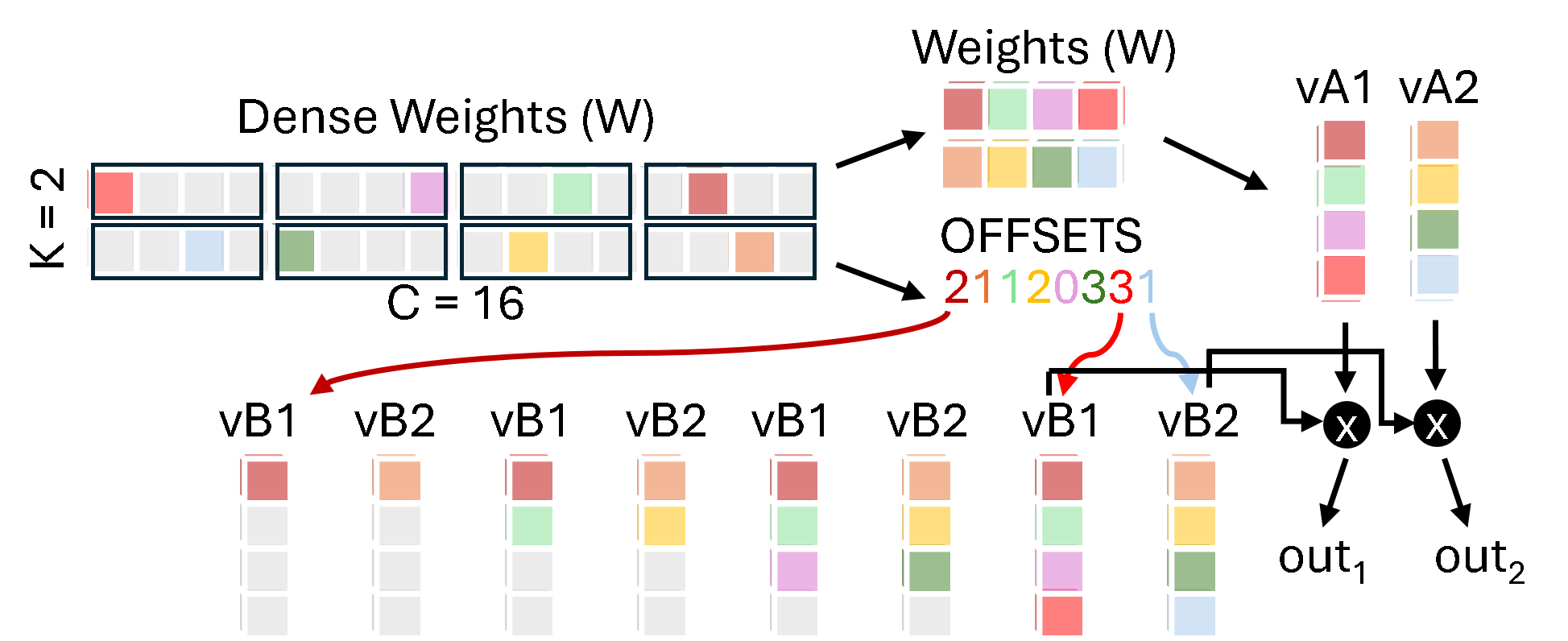}
    \caption{For FC layers, NZ \texttt{OFFSETS} are reorganized offline alternating 2 output channels. Activations are then loaded in two buffers \texttt{vB1} and \texttt{vB2} (example for 1:4 sparsity).}
    \label{fig:fc_isa_sort}
\end{figure}

The innermost loop of our ISA-extended sparse FC kernel is shown in Fig.~\ref{fig:fc_code} (right).
To make our HW extension lightweight, we exploit the same \texttt{xDecimate} instruction designed for Convolutions.
As mentioned in Sec.~\ref{subsec:conv_isa}, the instruction updates the M-sized block's base address, and the offset in \texttt{rd} only once every two executions.
For FC layers, which differently from Conv are not unrolled over two im2col buffers, we do not duplicate the NZ offsets, but rather, we reorganize them (offline).
Namely, we construct the \texttt{OFFSETS} array by alternating the offsets of NZ elements in two consecutive channels (i.e., $\mathtt{o}_{0,ch_i}$, $\mathtt{o}_{0,ch_{i+1}}$, $\mathtt{o}_{1,ch_i}$, $\mathtt{o}_{1,ch_{i+1}}$, etc.).
In the kernel's inner loop, we load NZ weights for channels $i$ and $i+1$ in two 4x8-bit registers (\texttt{vA1} and \texttt{vA2}), before calling \texttt{xDecimate} eight times to load 8 activations from a single im2col buffer, alternating \texttt{vB1} and \texttt{vB2} as destinations. As a result, \texttt{vB1} will eventually contain the activations corresponding to the $i$-th channel's NZ weights, and \texttt{vB2} those corresponding to the $i+1$-th channel, allowing to perform 8 MACs.
In other words, we essentially unroll over two output channels ($K=2$) despite not having input reuse opportunities, with the objective of maintaining the same exact instruction originally designed for Convolutions.
Fig.~\ref{fig:fc_isa_sort} shows the complete flow.
The theoretical peak performance obtained with this kernel is 0.61 dense equivalent MACs/instruction/core, which translates to 2.44, 4.88, and 9.76 MACs/instruction/core, always outperforming the dense baseline.

\subsection{xDecimate Hardware Implementation}\label{subsec:hw}
We prototyped the functionality of the \texttt{xDecimate} instruction at the register-transfer level (RTL) inside the open-source RI5CY/CV32E40P processor~\footnote{https://github.com/openhwgroup/cv32e40p} used in~\cite{vega}. The RTL is available open-source at \texttt{https://github.com/eml-eda/cv32e40x\_deci\\mate}.
RI5CY implements the \texttt{RV32IMC} ISA with the addition of the \texttt{Xpulpv2} extension, which includes hardware loops, load/store with post-increment, SIMD dot-product instructions, etc.
We selected a RISC-V processor for implementing our \texttt{xDecimate} instruction, leveraging the inherent extensibility of the ISA. However, in principle, our hardware extension could also be adapted to other architectures, such as ARM.

\begin{figure}[t]
    \centering
    \includegraphics[width=1.0\linewidth]{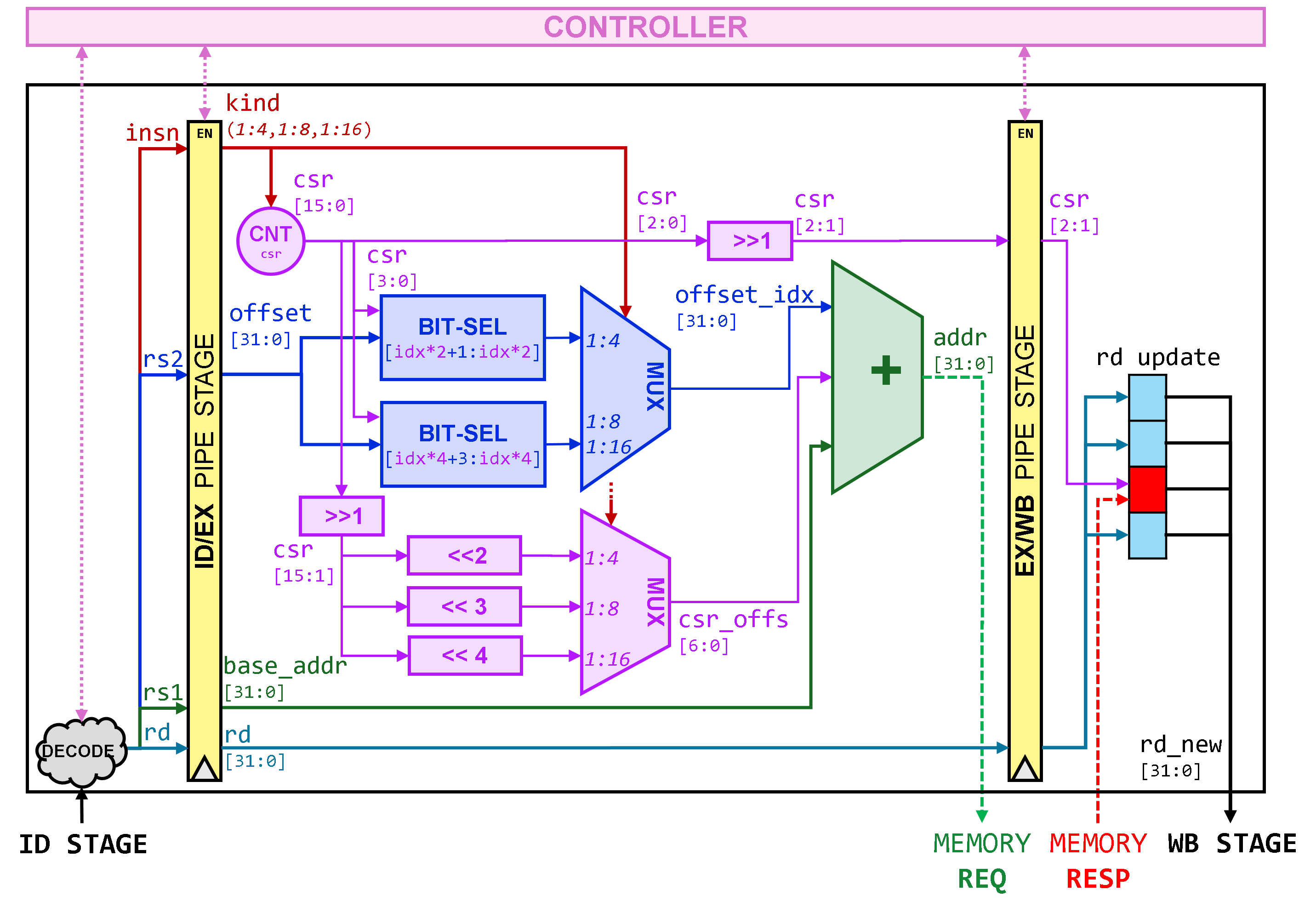}
    \vspace{-0.65cm}
    \caption{Detail of \texttt{xDecimate} eXtension Functional Unit (XFU) micro-architecture prototyped inside the RI5CY processor.}
    \label{fig:xifu_decimate}
\end{figure}

We modified the processor by adding an eXtension Functional Unit (XFU), whose microarchitecture is shown in
Fig.~\ref{fig:xifu_decimate}, which encompasses three of the four stages of RI5CY's pipeline: Instruction Decode (ID), Execute (EX), and Write-Back (WB).
Differently from regular RISC-V loads, the \texttt{xDecimate} instruction employs an R-type encoding with two source registers \texttt{rs1}, \texttt{rs2} and one destination register \texttt{rd}; all three registers are simultaneously read from the register file (RF, not shown in Fig.~\ref{fig:xifu_decimate}) in the ID stage, exploiting RI5CY's 3-read port RF, which is also necessary for the \texttt{Xpulpv2} extension.
A lightweight decoder identifies which flavor of \texttt{xDecimate} (i.e., which sparsity format) is being used.

The EX stage implements the extension's main functionality, which for 1:8 and 1:16 sparsity, can be summarized as:
\begin{align*}
&\mathtt{o} \leftarrow \mathtt{rs2[(csr[2:0]*4 + 3):(csr[2:0]*4)]}\\
&\mathtt{addr} \leftarrow \mathtt{rs1} + \mathtt{M} * \mathtt{csr[15:1]} + \mathtt{o}\\
\end{align*}
The \texttt{csr} is used to select the appropriate bits to unpack the NZ offset from the 32-bit word loaded in \texttt{rs2}.
The \texttt{csr} is also used to compute the starting address of the current M-sized block. For this, it is right-shifted by one position, thus ensuring that two consecutive \texttt{xDecimate} calls point to the same block (thus accounting for our inner loop unrolling). Both offsets are added to the base address loaded in \texttt{rs1}, and the result is propagated as a memory request through RI5CY's load/store unit. The instruction for 1:4 sparsity is similar, just using 4 \texttt{csr} LSBs instead of 3, given that \texttt{rs2} contains 16 2-bit wide offsets in this case.

In the WB stage, the interesting byte is extracted from the memory response and written into the \texttt{rd} register in the position selected by the \texttt{csr} (again right-shifted to account for unrolling). The updated \texttt{rd} is propagated back to RI5CY's RF. Then, the \texttt{csr} is automatically incremented to prepare for the next execution. Formally:
\begin{align*}
&\mathtt{rd[(csr[2:1]*8 + 7):(csr[2:1]*8)]} \leftarrow \mathtt{MEM[addr]}\\
&\mathtt{csr} \leftarrow \mathtt{csr} + 1\\
\end{align*}
The XFU controller also checks for data dependencies between consecutive \texttt{xDecimate} instructions to enable forwarding the value of \texttt{rd} from the WB stage.

\subsection{Integration in the MATCH Compiler}
\label{subsec:match}
To run end-to-end neural networks and support layer tiling, we integrate our sparse kernels into MATCH~\cite{match}, an extension to Apache TVM for heterogeneous SoCs. 
We added three main features to MATCH to support our sparse kernels efficiently:

\textbf{1) Modified Pattern Recognition:} the first compilation step of MATCH associates DNN graph patterns to known acceleration targets based on a HW-specific pattern table. Starting from the existing PULP target, we added new patterns that extend the ones already present for convolutional/FC layers, with an additional constraint on the weights' values, checking the NZ weights positions and recognizing 1:4, 1:8, or 1:16 sparsity formats.

\textbf{2) Tiling for Sparse Kernels:} the integration of N:M sparse kernels requires tiling both the NZ weight tensors and the associated NZ indices. Our modified tiling engine optimizes the tile layout by taking into account both the reduced dimension of the weights and the overhead due to the indices. To do so, we simply changed the number of bits associated with each weight. As an example, considering 1:4 sparsity, we need 12 bits to store each NZ weight (8 bits for the value, 4 bits for the replicated weight offset). Since the other 3 weights are zero, this is equivalent to having 3-bit per dense-equivalent weight.
Thanks to this modification to the tiling engine, we are able to tile sparse layers better, obtaining additional speed-ups due to better L1 utilization.

\textbf{3) Weights Memory Storage:} 
To further enhance performance, we modify the memory layout handling in MATCH to better support the simultaneous transfer of the compressed weights and their corresponding indices. In our approach, the weights and indices are stored in L2 memory in an interleaved fashion. Specifically, if a layer is tiled on $K/2$ output channels, MATCH stores the corresponding half of the weights, followed by the corresponding indices, so that both can be transferred with a single DMA transaction. Once the first tile has been processed, the second half of weights and indices are fetched in the same manner.

\begin{figure*}[t]
    \centering
    \includegraphics[width=0.99\linewidth]{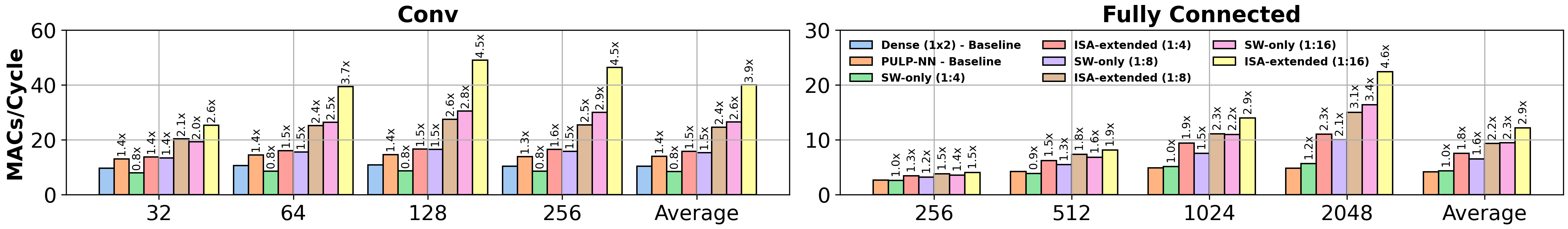}
    \caption{Single-layer results. Bars are grouped by input channels/features (C). Numbers report speedups over the dense 1x2 baseline.}
    \label{fig:results_layer_benchmarks}
    \vspace{-0.3cm}
\end{figure*}

\section{Results}
\begin{table*}[ht]
\footnotesize
\caption{End-to-end models results on the \cite{vega} platform.}\label{tab:deployment_results}
\begin{tabular}{llll|lll|lll}
         &          &             &       & \multicolumn{3}{c|}{\textbf{SW-only kernels}}         & \multicolumn{3}{c}{\textbf{ISA-extended kernels}}       \\\hline
\textbf{Model} &
  \textbf{Dataset} &
  \textbf{Sparsity} &
  \textbf{Acc.{[}\%{]}} &
  \textbf{MAC/cyc.} &
  \textbf{Cyc.[M]} &
  \textbf{Mem.{[}MB{]}} &
  \textbf{MAC/cyc.} &
  \textbf{Cyc.[M]} &
  \textbf{Mem.{[}MB{]}} \\ \hline\hline
\multirow{4}{*}{ViT}      & \multirow{4}{*}{CIFAR10}  & Dense & 95.59 & 4.65  & 975.23 & 21.59  &  -   &    -        &  -    \\
         &          & 1:4   & 95.73 & 4.80  & 944.17 & 11.86  & 6.66 & 681.19 & 11.86  \\
         &          & 1:8   & 95.02 & 6.31  & 718.86 & 10.09  & 7.48 & 606.99 & 10.09 \\
         &          & 1:16  & 95.17 & 7.59  & 598.04 & 8.76   & 8.40 & 540.23 & 8.76 \\ \hline
\multirow{4}{*}{ResNet18} & \multirow{4}{*}{CIFAR100} & Dense 1x2 & 75.28 & 8.33 & 66.63 & 11.22    &  -   &   -     &  -      \\
         &          & PULP-NN & 75.28 & 11.17 & 49.71 & 11.22   &   -  &   -     &  -      \\
         &          & 1:4   & 75.78 & 8.11  & 68.44 &  3.66   & 14.74 & 37.67  & 4.35 \\
         &          & 1:8   & 75.63 & 14.78 & 37.57 &  2.29   & 23.12 & 24.01  & 2.98 \\
         &          & 1:16  & 73.79 & 25.85 & 21.48 &  1.26     & 35.87 & 15.48  & 1.6\\\hline
\end{tabular}
\vspace{-0.5cm}
\end{table*}

\begin{table}[ht]
\footnotesize
\centering
\begin{threeparttable}[b]
\caption{Comparison with the state-of-the-art.}\label{tab:sota}
\centering
\begin{tabular}{l|lll}
\textbf{Benchmark}          &  \textbf{Spars.} & \textbf{Speedup} & \textbf{Area[\%]} \\ \hline\hline
LeNet \tnote{1} &  93.28\%                 & 3.51   &    -      \\
ConvNet \tnote{1} &  59.9\%                 & 1.38   &    -      \\
LeNet300 \tnote{1} &  93.07\%                 & 9.17   &    -      \\
DS-CNN \tnote{2}     &  90\%                    & 1.71  &    -        \\
ResNet50\tnote{3}        &  75\%                  & 1.82$^*$   &    n.a.       \\
DenseNet\tnote{3}        &  75\%                  & 2.14$^*$  &    n.a.       \\
InceptionV3\tnote{3}        &  75\%                  & 1.92$^*$   &    n.a.       \\
spMV\tnote{4}        &  95.7\%                  & 5$^*$   &    44       \\

\textbf{ResNet18-SW}               &  \textbf{87.5-93.75\%}         & \textbf{1.77-3.10} &    -         \\
\textbf{ResNet18-ISA}               &  \textbf{75-93.75\%}         & \textbf{1.77-4.31} &    \textbf{5}        \\\hline
\end{tabular}
\begin{tablenotes}
\item [1] \cite{scalpel}, $^2$ \cite{dcsr}, $^3$ \cite{indexmac}, $^4$ \cite{sssr}, $^*$ speedup compared to SW-only sparse baseline
\end{tablenotes}
\end{threeparttable}
\vspace{-0.6cm}
\end{table}

\subsection{Experimental Setup}
We benchmark our proposed kernels and two end-to-end networks using the GVSoC virtual platform~\cite{gvsoc} to simulate Vega \cite{vega}.
The HW extension has been implemented in SystemVerilog RTL and synthesized using Synopsys Design Compiler targeting the same 22nm technology as~\cite{vega} with \SI{200}{MHz} target clock frequency in worst-case operating conditions (slow-slow process, \SI{0.72}{\volt} operating voltage, \SI{125}{C} temperature).
The two DNNs are a ResNet18~\cite{resnet}, trained on Cifar100, and a Vision Transformer~\cite{vit} (ViT-Small), trained on a rescaled 224x224 version of Cifar10. Both have been trained for 200 epochs using the combined training and pruning scheme detailed in~\cite{nm_mask_training}.
For ResNet, we apply N:M pruning to 3x3 convolutions, leaving pointwise layers dense. For the ViT, we sparsify only the FC layers of the feed-forward block. Note that these layers are part of many other relevant transformer-based architectures, such as~\cite{bert,t5}. As a consequence, our approach is easily transferable to other architectures, where comparable savings to our benchmarks on ViT can be expected.
Note also that both the task and the training of the pruned networks are orthogonal to our work, making our approach usable also on more complex datasets.

After training, we quantize all networks to 8 bits using Brevitas~\cite{brevitas}.
All individual kernels and end-to-end ResNets have been deployed using the modified MATCH compiler. For ViTs, given the lack of support for attention layers in MATCH, we computed the latency layer-by-layer, using Deeploy \cite{deeploy}, an alternative compiler for the same HW, for attention layers, and MATCH for sparse/dense feed-forward layers.
The reported MACs/cycle values always refer to dense-equivalent operations.

\subsection{Single Layers Benchmarking}
We first benchmark both convolutional and FC layers, varying the input channels/neurons dimension while keeping other parameters fixed; we set the output channels/neurons K = 256, and for convolutions, we also use IX/IY = OX/OY = 8, FX/FY = 3, S = 1, P = 1. We vary C $\in$ [32, 64, 128, 256] for convolutions and C $\in$ [256, 512, 1024, 2048] for FC layers.
Fig.~\ref{fig:results_layer_benchmarks} shows the performance of 1:4, 1:8, and 1:16 sparse kernels, without (SW) our with (ISA) \texttt{xDecimate}, as well as the results of the two dense baselines, 1x2 and PULP-NN (FC layers have a single baseline, given that also PULP-NN uses 1x2 unrolling).

As expected from the inner loop instructions analysis of Sec.~\ref{sec:methods} the 1:4 SW-only Convolution performs worse than the dense versions (+23\% cycles on average, w.r.t. the 1x2 baseline). At the other extreme, the 1:16 SW-only kernel obtains an average speedup of 2.6$\times$ and 1.85$\times$ compared to the 1x2 and the PULP-NN baselines. Notice that the lower speed-up compared to the theoretical one obtained comparing inner loop instructions (3.6$\times$ vs the 1x2 baseline) is mainly due to the im2col step, which is identical in sparse and dense kernels and reduces the overall MACs/cycle performance. 
ISA-extended kernels further improve performance, making 1:4 sparsity advantageous too, with an average speedup of 1.50$\times$ and 1.12$\times$ over 1x2 and PULP-NN. At 1:8 sparsity, the speedups grow to 2.4$\times$ and 1.74$\times$, and at 1:16 sparsity, to 3.9$\times$ and 2.78$\times$, respectively.
For all kernels, performance tends to improve with increasing C; the improvement is more marked for sparser layers, given that our kernels accelerate the innermost matrix multiplication loop, whose relative importance over the constant im2col phase grows with C. At C=256, given the high occupation of the im2col buffer, L1 tiles for Conv layers over spatial dimensions start to be very small. Thus, performance reduces slightly.

For FC layers, SW-only sparse kernels outperform the baselines even at 1:4 sparsity, albeit with modest latency reductions (2\% on average).
This is particularly evident on larger geometries (e.g., C = 2048), where the speedup reaches 1.2$\times$.
Although there is no theoretical gain in the inner loop, this improvement derives from the fewer weight loads required, given the lower memory footprint of sparse weights. This contribution is less evident (or absent) for convolutions, where the latency of weights transfers from L2 to L1 is hidden using double-buffering \cite{match}. For memory-bound FC layers, instead, these transfers are one of the dominant components of the overall latency.
At higher sparsity ratios (1:8 and 1:16), the average speedups increase to 1.6$\times$ and 2.3$\times$, respectively, with peaks up to 3.4$\times$.
ISA-extended kernels yield more significant improvements, achieving an average 1.8$\times$ speedup at 1:4 sparsity with a peak of 2.3$\times$ for C = 2048. At 1:8 and 1:16 sparsity, the average speedups are 2.2$\times$ and 2.9$\times$, respectively.

\subsection{End-to-end DNNs Benchmarking}
Table~\ref{tab:deployment_results} reports profiling results of end-to-end models. We also report the accuracy obtained by our sparse models to demonstrate that the considered sparsity patterns are applicable to real-world applications, leading in most cases to null or small degradations.

In the ViT model, the sparsified FC layers account for 65\% of the model's parameters and 60\% of the operations.
Despite the relevant parameters reduction, even with the most aggressive pruning pattern (1:16), sparsifying these layers impacts the accuracy minimally, with a drop of 0.42\%.
On the other hand, in terms of latency, all sparsified models, both with and without the new \texttt{xDecimate} operation, outperform the dense baseline.
Using SW-only kernels, we achieve speedups of 1.03$\times$, 1.36$\times$, and 1.63$\times$, respectively, at 1:4, 1:8, and 1:16 sparsity.
ISA-extended kernels outperform the dense models even further, achieving speedups of 1.43$\times$, 1.61$\times$, and 1.81$\times$. 
Overall, our most compressed model (1:16) achieves 95.17\% accuracy with a 2.34$\times$ lower memory footprint compared to the dense counterpart and a 1.81$\times$ latency reduction when using our ISA extension.

Concerning ResNet18, the sparsified convolutions (all but the pointwise) account for 97\% of the total parameters and 98\% of the total MACs.
Training the models with 1:4 and 1:8 sparsity, we obtained even higher accuracy than the dense network, while 1:16 yields a slight performance drop of 1.49\%.
In terms of latency, similarly to the result obtained on single layers, the model that leverages the 1:4 SW-only kernels is outperformed by both dense baselines (1x2 and PULP-NN), achieving 1.03$\times$ and 1.38$\times$ higher latency, although it reduces the memory footprint from 11.22 MB to 3.66 MB.
SW-only sparse models outperform their dense counterparts at 1:8 and 1:16 sparsity.
With the ISA extension, all sparse ResNets achieve speedups against both baselines.
The most compressed model that does not incur accuracy drops uses 1:8 sparsity, reaching 0.35\% higher accuracy, 2.07$\times$ lower latency, and 3.77$\times$ lower memory footprint compared to the best dense option, i.e., PULP-NN.
Notice that ResNet versions using \texttt{xDecimate} require slightly more memory than SW-only ones due to the replication of NZ indices in Conv kernels (see Sec.~\ref{subsec:conv_isa}).

While our benchmark DNNs are relatively small, compatibly with edge applications' constraints, we underline that even greater speedups would be achieved on bigger models, given the combined effect of a more aggressive pruning applicable to them and the lower impact of overhead effects (e.g., external loop management) on bigger layers, in which most of the complexity resides in the internal convolutional kernel loops that we optimize.

\subsection{SoTA Comparison}
Table~\ref{tab:sota} compares our work with the SotA on sparse acceleration for MCUs. 
We report the benchmark DNNs, the sparsity level, the speedup obtained w.r.t. a dense execution, and the area overhead, if present. 
Concerning our results, we report those on ResNet since, to our knowledge, no SotA work considered N:M pruning for transformers on MCUs.

The authors of~\cite{scalpel} report speedups from 1.31$\times$ to 9.17$\times$.
At a high sparsity ($\geq$90\%), they achieve 3.21$\times$ speed-up over a dense LeNet CNN and 9.17$\times$ on the FC-only LeNet300. As anticipated in Sec.~\ref{sec:related}, the high speed-up on the LeNet300 is caused by the high-latency loads from Flash memory. Since FC models are memory-bound, strongly reducing the number of loads almost linearly translates to latency savings.
Conversely, on more compute-bound CNNs, the load latency impact is lower, and our ISA-extended solution achieves better speed-ups for both aggressive (4.31$\times$ vs 3.51$\times$ at $\geq$90\% sparsity) and moderate pruning (1.77$\times$ vs 1.38$\times$ for 75\% vs 59.9\% sparsity). With SW-only kernels, we achieve comparable results.

The Large DS-CNN from~\cite{dcsr}, at 90\% sparsity, achieves 1.71$\times$ speed-up compared to dense unoptimized CMSIS-NN kernels.
Similarly, at 87.5\% sparsity, we obtain 1.77$\times$/2.77$\times$ speed-ups with the SW and ISA kernels compared to the 1x2 baseline.
Note also that, with the N:M format, we obtain a memory reduction of 79.59/73.44\%, compared to the 74.45\% obtained by~\cite{dcsr}, with similar levels of sparsity (87.5\% vs 90\%).

In~\cite{indexmac}, the authors report speedups of 1.82/1.92/2.14$\times$ on different convolutional networks at 75\% sparsity, comparing their ISA-enhanced results to SW-only sparse kernels.
On our ResNet18 model at iso-sparsity, thanks to \texttt{xDecimate}, we achieve a 1.82$\times$ speedup w.r.t. the SW-only version, despite the fact that our HW extension, being tailored for ultra-low-power edge devices, is much more lightweight than the one in~\cite{indexmac}, which leverages a large vector RF on a more powerful core (and whose area overhead is not reported).

In~\cite{sssr}, a 5$\times$ speedup over SW-only sparse kernels for GEMM is reported, at 95.7\% sparsity, higher than the 1.39$\times$@93.75\% sparsity (i.e. 1:16) that we achieve with our ISA extension vs the SW-only ResNet18. However, note that for such extreme values a sparsity, an apparently small difference corresponds to executing 6.25\% of the original MACs (in our case), vs 4.3\% (i.e. 1.45$\times$ fewer) in \cite{sssr}.
Furthermore, the HW support required to achieve such an acceleration leads to an area overhead of 44\% w.r.t. an FPU-less core like ours, versus our 5\%.

\section{Conclusions}
We have introduced a set of efficient kernels N:M pruning on MCUs of the PULP family, and a lightweight extension of the XPulpV2 ISA, to further speed them up. On the platform described in~\cite{vega}, and targeting a ViT and a ResNet18, we have achieved end-to-end latency reductions of 1.81$\times$ and 3.21$\times$, with less than 0.5/1.5\% accuracy drop on CIFAR10/CIFAR100 respectively.
Our future work will study the impact of variable sparsity patterns (e.g., per-layer or per-channel) on latency and accuracy, considering both the pure software kernels and the \texttt{xDecimate}-enhanced ones. Further, we will prototype our hardware extension on FPGA to enable an estimation of the energy savings achieved by our kernels, which can show further advantages in the reduced off-chip memory accesses.

\end{document}